\documentclass[letterpaper]{article} % DO NOT CHANGE THIS
\usepackage{aaai2026}  % DO NOT CHANGE THIS
\usepackage{times}  % DO NOT CHANGE THIS
\usepackage{helvet}  % DO NOT CHANGE THIS
\usepackage{courier}  % DO NOT CHANGE THIS
\usepackage[hyphens]{url}  % DO NOT CHANGE THIS
\usepackage{graphicx} % DO NOT CHANGE THIS
\usepackage{natbib}  % DO NOT CHANGE THIS AND DO NOT ADD ANY OPTIONS TO IT
\usepackage{caption} % DO NOT CHANGE THIS AND DO NOT ADD ANY OPTIONS TO IT
\usepackage{subcaption}
\usepackage{algorithm}
\usepackage{algorithmic}
\usepackage{amsmath}
\usepackage{enumitem}
\usepackage{booktabs}
\usepackage{multirow}
\newcommand{\tabincell}[2]{\begin{tabular}{@{}#1@{}}#2\end{tabular}}
\usepackage{newfloat}
\usepackage{listings}
\DeclareCaptionStyle{ruled}{labelfont=normalfont,labelsep=colon,strut=off} % DO NOT CHANGE THIS
\floatstyle{ruled}
\newfloat{listing}{tb}{lst}{}
\floatname{listing}{Listing}
\title{GreatSplicing: A Semantically Rich Splicing Dataset}
\author {Jiaming Liang, Yuwan Xue, Haowei Liu, Zhenqi Dai, Yu Liao, Rui Wang,\\ Weihao Jiang, Yaping Liu, Zhikun Chen, Guoxiao Liu, Bo Liu, Xiuli Bi\footnote{Corresponding author.}}
\affiliations {Chongqing Key Laboratory of Image Cognition, Chongqing University of Posts and Telecommunications, China\\
chinaliangjm@gmail.com, xueywuestc@gmail.com, S230201069@stu.cqupt.edu.cn, 2991541345@qq.com, s200201073@stu.cqupt.edu.cn, 18182241057@163.com, weihaojiang@foxmail.com, 1942169033@qq.com, cheeko0n@163.com, 1872828971@qq.com, boliu@cqupt.edu.cn, bixl@cqupt.edu.cn
}
\begin{document}

\maketitle

\begin{abstract}
In existing splicing forgery datasets, the insufficient semantic variety of spliced regions causes trained detection models to overfit semantic features rather than learn genuine splicing traces. Meanwhile, the lack of a reasonable benchmark dataset has led to inconsistent experimental settings across existing detection methods. To address these issues, we propose GreatSplicing, a manually created, large-scale, high-quality splicing dataset. GreatSplicing comprises 5,000 spliced images and covers spliced regions across 335 distinct semantic categories, enabling detection models to learn splicing traces more effectively. Empirical results show that detection models trained on GreatSplicing achieve low misidentification rates and stronger cross-dataset generalization compared to existing datasets. GreatSplicing is now publicly available for research purposes at the following link.
\end{abstract}

\begin{links}
    \link{Dataset}{http://www.greatsplicing.net/}
\end{links}

\section{Introduction}

In recent years, image forensics has become increasingly critical. Besides Deepfake and other generated images, traditional image forgery, which alters the local semantic information of an image while maintaining its overall authenticity and integrity, is still often exploited by malicious users. Among them, splicing forgery is one of the most common local forgeries\cite{bi2023self}, which is defined\cite{ng2004model} as the cut-and-paste of image regions from one image onto another image to create a spliced image. Splicing forgery detection\cite{ahmad2021digital} is to localize the spliced regions and disclose them via a binary map.

\begin{figure}[t]
    \centering
    \includegraphics[scale=0.18]{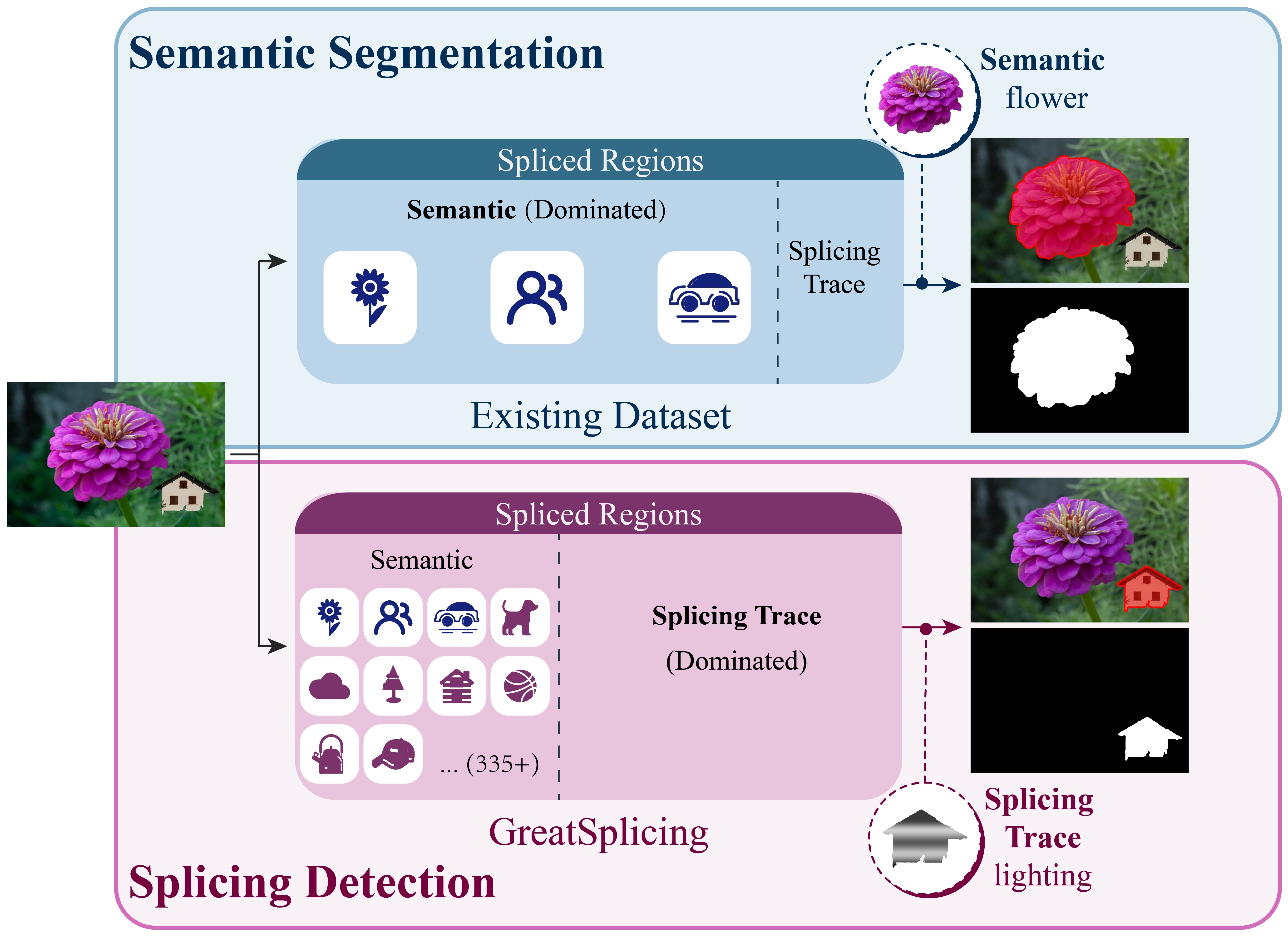}
    \caption{The distinctions between GreatSplicing and existing splicing datasets. The spliced regions in GreatSplicing exhibit a wider variety of semantic categories, which enhances the network's learning of splicing traces.}
    \label{figure_distinction_greatsplicing_and_existing_datasets}
\end{figure}

Thanks to the powerful pattern recognition capabilities of neural networks, numerous detection models for splicing forgery have been proposed. However, subject to the limitations of existing datasets, these models cannot be fully exploited and perform unsatisfactorily in cross-dataset cases. Meanwhile, due to the absence of a dataset that can faithfully reflect the performance of these methods, the experimental settings in different works are often based on independently synthesized datasets for training. During the training-testing process, specific techniques like image augmentation\cite{shorten2019survey} and fine-tuning are employed. It raises great concerns about fairness.

Therefore, to address the problems caused by the absence of well-structured splicing datasets, we propose GreatSplicing. Based on BossBase, GreatSplicing comprises 5,000 spliced images that were manually created with Adobe Photoshop, without any automation. \textbf{The spliced regions in GreatSplicing exhibit substantial semantic diversity, which prevents detection models from overfitting to specific semantics and allows them to focus on splicing traces, thereby improving performance.} GreatSplicing features high resolution and follows a standardized production process, ensuring the faithful preservation of splicing traces. The distinctions between GreatSplicing and existing datasets are illustrated in Figure \ref{figure_distinction_greatsplicing_and_existing_datasets}.

The main contributions are summarized as follows:
\begin{itemize}

\item To the best of our knowledge, this is the first work to reveal the significant defect of insufficient semantic diversity in existing splicing forgery datasets. Empirical results demonstrate that this limitation can result in severe misidentification.

\item We propose GreatSplicing, a manually created, semantically rich splicing dataset to address this challenge. GreatSplicing features high resolution and follows a standardized production process, ensuring the faithful preservation of splicing traces.

\item Extensive empirical results demonstrate that, compared with existing splicing forgery datasets, GreatSplicing has significant advantages in preventing semantic overfitting and in extracting genuine splicing traces.
\end{itemize}

The rest of the paper is organized as follows. In Section 2, a detailed introduction to existing splicing forgery datasets will be provided. In addition, an overview of BossBase will be presented. In Section 3, we will present the production process of GreatSplicing. Subsequently, Section 4 will present extensive empirical results to verify the superiority of GreatSplicing. Finally, Section 5 will summarize this work.

\section{Related Works}
This section will introduce existing splicing forgery datasets and their limitations. Besides, information about the authentic image dataset BossBase will be provided.

\subsection{Existing Splicing Datasets}

\noindent\textbf{CASIA}\cite{dong2013casia} consists of V1.0 and V2.0, comprising 8,000 authentic color images and 2,300 spliced images. The authentic images used in creating CASIA encompass nine different scenes.

\noindent\textbf{DEFACTO}\cite{mahfoudi2019defacto} is a large-scale image tampering dataset comprising 105,000 spliced images. However, the spliced regions cover only seven semantic categories: airplane, bird, clock, frisbee, ball, traffic light, and stop sign.

\noindent\textbf{NIST16}\cite{guan2019mfc} comprises 288 spliced images, with over half of the images featuring combinations of basic shapes, such as spheres, triangles, and cubes.

\noindent\textbf{FantasticReality}\cite{kniaz2019point} consists of 16,592 authentic images and 14,546 spliced images. These authentic images were collected from undisclosed sources and employed to create the spliced images. Besides, the spliced regions in FantasticReality encompass only ten semantic categories: person, car, truck, van, bus, building, cat, dog, tram, and boat.

\noindent\textbf{Columbia}\cite{hsu2006columbia} only contains 180 spliced images, and all spliced regions in these images are of random-shapes.

\noindent\textbf{IMD2020}\cite{novozamsky2020imd2020} offers 2,010 spliced images and their corresponding authentic images. However, these spliced images are derived from the Internet and created by random individuals. Additionally, these spliced images have undergone post-processing, such as JPEG compression. Therefore, there are significant concerns about the rigor of the production standards.

\begin{table}[t]
    \centering
    \begin{subtable}[t]{1\linewidth}
        \centering
        \fontsize{9}{10}\selectfont
        \setlength\tabcolsep{4.5pt}
        \renewcommand{\arraystretch}{1.2}
        \begin{tabular}{c@{\hspace{1pt}}l@{\hspace{1pt}}l@{\hspace{1pt}}l@{\hspace{1pt}}c@{\hspace{1pt}}c}
            \toprule
            Order & \multicolumn{1}{l}{Datasets} & \multicolumn{1}{l}{Capacity} &   \multicolumn{1}{c}{Semantics} &O &R\\
            \midrule
            1 & CASIA V2.0 & \multicolumn{1}{l}{$1,849$} & \multicolumn{1}{c}{\tabincell{l}{\textless 30}}& \multicolumn{1}{c}{$\bullet$} &\\
            
            2 & FantasticReality & \multicolumn{1}{l}{$14,546$} & \multicolumn{1}{c}{\tabincell{l}{= 10}}& \multicolumn{1}{c}{$\bullet$} &\\
            
            3 & DEFACTO & \multicolumn{1}{l}{$105,000$} & \multicolumn{1}{c}{\tabincell{l}{= 7}}& \multicolumn{1}{c}{$\bullet$} &\\
            
            4 & NIST16 & \multicolumn{1}{l}{$288$} & \multicolumn{1}{c}{\tabincell{l}{\textless 20}}& \multicolumn{1}{c}{$\bullet$} &\\
            
            5 & Columbia & \multicolumn{1}{l}{$180$} & \multicolumn{1}{c}{\tabincell{l}{N/A}}& &\multicolumn{1}{c}{$\bullet$}\\
            
            6 & IMD2020 &\multicolumn{1}{l}{$2,010$}   & \multicolumn{1}{c}{\tabincell{l}{N/A}}& \multicolumn{1}{c}{$\bullet$} &\\
            
            7 & GreatSplicing & \multicolumn{1}{l}{$5,000$}   & \multicolumn{1}{c}{\tabincell{l}{=335}} & \multicolumn{1}{c}{$\bullet$} & \multicolumn{1}{c}{$\bullet$}\\
            
            \bottomrule
        \end{tabular}
        \caption{Summary of existing public splicing datasets. "Semantics" column reports a rough estimate of the number of semantic categories of spliced regions contrained in each dataset. "O" and "R" denote the inclusion of object-aware and shape-random spliced images, respectively.}
        
    \end{subtable}
    \begin{subtable}[t]{1\linewidth}
        \centering
        \fontsize{9}{10}\selectfont
	\setlength\tabcolsep{4.5pt}
        \renewcommand{\arraystretch}{1.2}
        \begin{tabular}{c@{\hspace{4pt}}l@{\hspace{4pt}}l@{\hspace{4pt}}c@{\hspace{4pt}}c@{\hspace{4pt}}c@{\hspace{4pt}}c}
            \toprule
            Time &Model & \multicolumn{1}{l}{Dataset}& Syn & Fine & Aug & Auth  \\
            \midrule
            \multicolumn{1}{c}{2022} &MVSS-Net     &\multicolumn{1}{c}{\tabincell{l}{1, 3, 4, 5, 6}}&   & & &  \multicolumn{1}{c}{$\bullet$}     \\
            
            \multicolumn{1}{c}{2022} &Objectformer     & \multicolumn{1}{l}{1, 4, 5, 6}&  \multicolumn{1}{c}{$\bullet$}    & \multicolumn{1}{c}{$\bullet$}    & \multicolumn{1}{c}{$\bullet$}    &  \multicolumn{1}{c}{$\bullet$}   \\
            \multicolumn{1}{c}{2022} &PSCC-Net    &\multicolumn{1}{l}{1, 4, 5, 6}&   \multicolumn{1}{c}{$\bullet$}    &     & \multicolumn{1}{c}{$\bullet$} &\\
            \multicolumn{1}{c}{2021} &RTAG& \multicolumn{1}{l}{1, 2, 4, 5}     &    & &  & \\
            
            \multicolumn{1}{c}{2020} &SPAN  & \multicolumn{1}{l}{1, 4, 5}   &  \multicolumn{1}{c}{$\bullet$}    & \multicolumn{1}{c}{$\bullet$}    &   &   \\
            
            \multicolumn{1}{c}{2019} &RRU-Net & \multicolumn{1}{l}{1, 5}    &  && \multicolumn{1}{c}{$\bullet$}    &  \\
            
             \multicolumn{1}{c}{2019} &Mantra-Net  & \multicolumn{1}{l}{1, 5, 6}      & \multicolumn{1}{c}{$\bullet$}    &  &  & \multicolumn{1}{c}{$\bullet$}   \\
            \multicolumn{1}{c}{2019}&MAGritte &\multicolumn{1}{l}{1, 2, 5}&&&&\multicolumn{1}{c}{$\bullet$}\\
             \multicolumn{1}{c}{2019} &HLED&\multicolumn{1}{l}{4} & \multicolumn{1}{c}{$\bullet$}&\multicolumn{1}{c}{$\bullet$}& \multicolumn{1}{c}{$\bullet$}    &   \\
            \multicolumn{1}{c}{2018} &RGB-Noise&\multicolumn{1}{l}{1, 4, 5}     & \multicolumn{1}{c}{$\bullet$}    & \multicolumn{1}{c}{$\bullet$}    &  &  \\
             \multicolumn{1}{c}{2017} &ESS-Net &\multicolumn{1}{l}{4}    &  &  & \multicolumn{1}{c}{$\bullet$}     &  \\
            \bottomrule
           
        \end{tabular}
        \caption{Summary of various experimental settings over the past six years. "Dataset" column reports the datasets used in each work, with the numbers corresponding to the dataset indices in Table (a). "Syn" denotes the use of synthetic datasets for training. "Fine" refers to fine-tuning. "Aug" and "Auth" represent the inclusion of augmented and authentic images during training, respectively.}
    \end{subtable}
    \caption{\normalsize Summary of datasets and experimental settings in the current splicing detection field.}
    \label{table_summary_datasets_and_settings}
\end{table}

\begin{figure*}[t]
    \centering
    \includegraphics[scale=0.52]{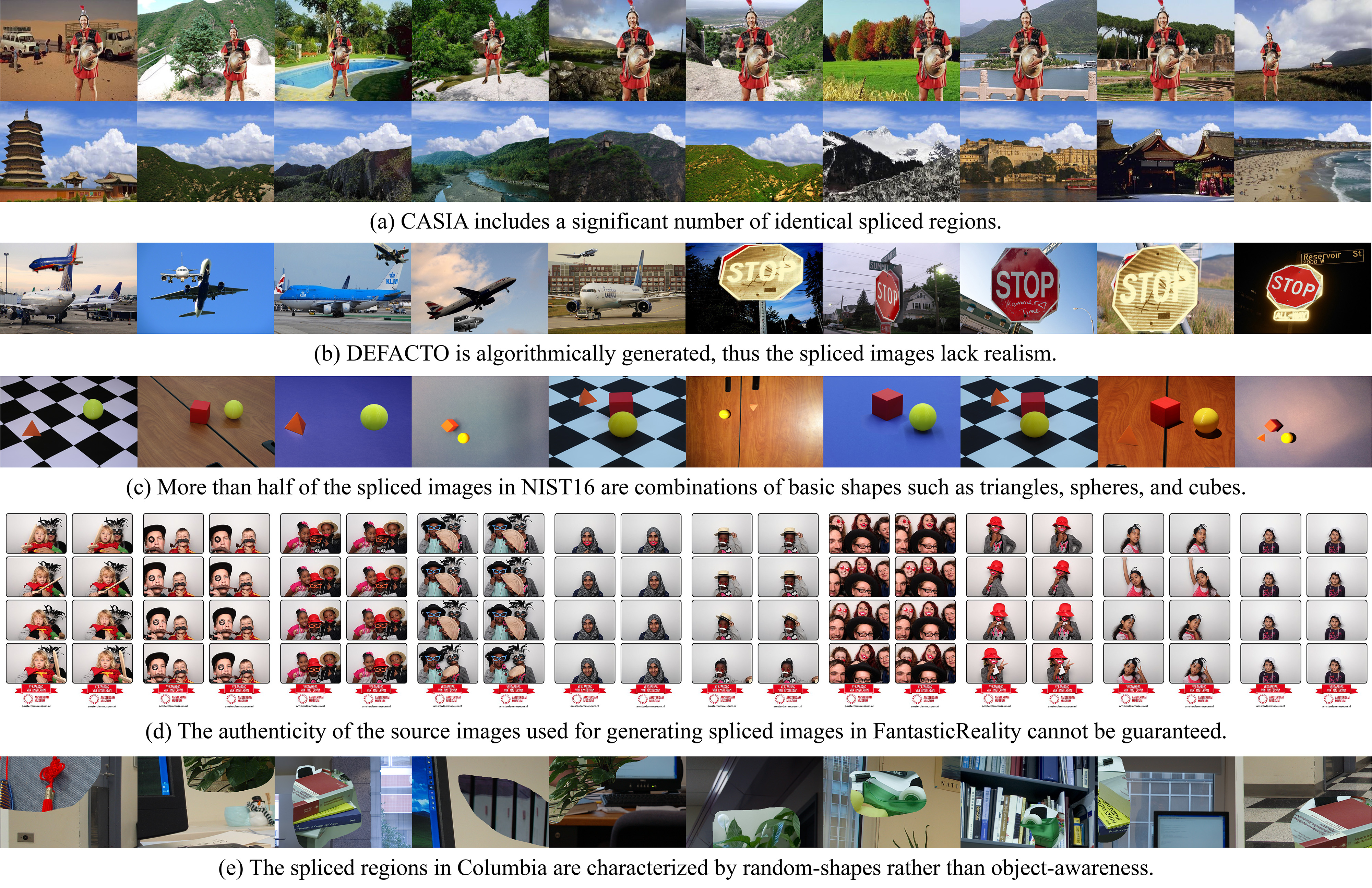}
    \caption{Examples of limitations in existing splicing datasets.}
    \label{figure_limitations_in_existing_datasets}
\end{figure*}

\subsection{BossBase} 
Our GreatSplicing is based on BossBase\cite{bas2011break} dataset. BossBase is a widely used dataset in image steganalysis, comprising 10,000 images taken with digital cameras and stored in their original camera image format. BossBase encompasses images from 7 distinct camera models, namely the Canon EOS 400D, Canon EOS 40D, Canon EOS 7D, Canon EOS DIGITAL REBEL XSi, PENTAX K20D, NIKON D70, and M9 Digital Camera, covering diverse scenes from daily life. The images are saved in multiple formats, including CR2, PEF, DNG, and NEF, and they exhibit high resolution and considerable size. BossBase was selected to create GreatSplicing for the following reasons:
\begin{enumerate}
    \item BossBase ensures image authenticity because it originates from cameras;
    \item BossBase covers diverse real-world scenarios, enabling the creation of spliced images that are better aligned with practical applications;
    \item The abundant semantic information present in these real-world images provides an advantage for generating a splicing detection dataset with rich semantic content;
    \item The high resolution of the images in BossBase can help us preserve splicing traces effectively.
\end{enumerate}

\section{GreatSplicing}

\subsection{Limitations of Existing Splicing Forgery Datasets}
While various splicing forgery datasets are available, the less rigorous creation processes have introduced several significant deficiencies. Consequently, existing detection models cannot fully realize their potential. 

For example: (1) \textbf{CASIA} includes numerous identical spliced regions, as shown in Figure~\ref{figure_limitations_in_existing_datasets}(a). Spliced regions in the first row are the same people, and in the second row, the skies are the same. Similar instances are considerable in CASIA, leading models to overfit identical semantic information, thereby impeding the learning of splicing traces. (2) \textbf{DEFACTO} is algorithmically generated based on MSCOCO\cite{lin2014microsoft}, as illustrated in Figure \ref{figure_limitations_in_existing_datasets}(b). These spliced images lack verisimilitude and do not align with our real world. Thus, detection models trained on DEFACTO are difficult to apply. (3) More than half of the spliced regions in \textbf{NIST16} consist of basic geometric shapes, as depicted in Figure \ref{figure_limitations_in_existing_datasets}(c). Detection Models trained on NIST16 tend to overfit these simple shapes rather than splicing traces. (4) The authenticity of the authentic images employed for generating spliced images in \textbf{FantasticReality} cannot be guaranteed, as illustrated in Figure \ref{figure_limitations_in_existing_datasets}(d). Each of these ten images is a splice of smaller images. However, they are annotated as "authentic" in FantasticReality. This error will confuse the detection model regarding splicing traces. (5) Spliced regions within \textbf{Columbia} exhibit random shapes, as illustrated in Figure \ref{figure_limitations_in_existing_datasets}(e). These images exhibit noticeable edge artifacts that cause detection models to overfit to them rather than learning comprehensive splicing traces, unlike our practically splicied images.

In addition to the aforementioned concerns, a prevalent issue in existing splicing datasets is the limited semantic diversity of spliced regions. Table~\ref{table_summary_datasets_and_settings}(a) summarizes the number of semantic categories within the spliced regions of each dataset. Consequently, trained detection models tend to become overly sensitive and overfit to specific semantics, which hinders their ability to learn effective splicing traces and often results in severe misidentification and reduced detection performance.

In summary, existing datasets in the field of splicing forgery detection exhibit various limitations. To address this challenge, we propose GreatSplicing as a solution.

\subsection{GreatSplicing}
For ease of describing GreatSplicing, the following definitions are stated briefly. 

A spliced image is defined as
\begin{equation}
    s := x_{f} \odot M + x_{b} \odot (1 - M).
    \label{defination_spliced_image}
\end{equation}
Here, $x_f$ and $x_b$ refer to two different authentic image sources, while $M$ represents a binary image mask where each element takes value from \{0, 1\}, with the symbol $\odot$ indicating element-wise multiplication. It should be pointed out that
\begin{equation}
    \sum_{(i,j)}M(i,j) < \sum_{(i,j)}[1-M(i,j)].
    \label{defination_front_region_area}
\end{equation}
This inequation suggests that the area of the spliced region expressed by $x_{f} \odot M$ should be smaller than the area of the background region expressed by $x_{b} \odot (1 - M)$.

\begin{table}
    \centering
    \fontsize{9}{10}\selectfont
     \setlength\tabcolsep{5.0pt}
    \renewcommand{\arraystretch}{1.2}
    \begin{tabular}{lc}
        \toprule
        Category & Details\\
        \midrule
        \textbf{Cut-Out Tools} &\tabincell{c}{1) marquee tool, 2) lasso tool,\\3) magic wand tool, 4) pen tool,\\5) quick selection tool} \\
        \midrule
        \textbf{Image Manipulations} &\tabincell{c}{1) scale, 2) rotate, 3) skew,\\4) distort, 5) brightness/contrast,\\6) flip, 7)levels, 8) hue/saturation,\\9) curves, 10) color balance}\\
        \bottomrule
    \end{tabular}
    \caption{Summary of the tools and image manipulations used in creating GreatSplicing.}
    \label{summary_photoshop}
\end{table}

The process is meticulously controlled while creating GreatSplicing to avoid shortcomings present in existing splicing datasets. It can be summarized as
\begin{equation}
    M \odot (\Pi_{i=1}^{k_{f}}\varphi_{i})(x_{f}) + (1 - M) \odot (\Pi_{j=1}^{k_{b}}\phi_{j})(x_{b}) \rightarrow s.
    \label{make_GreatSplicing}
\end{equation}
Specifically, the process involves these four steps:

\begin{enumerate}
\item Selecting two images, $x_f$ and $x_b$, from BossBase, to be used for the spliced region source and the background region source respectively. Each image in BossBase is used only once, creating 5,000 spliced images from 10,000 authentic images. 

\item Applying one or more of the image manipulations shown in Table \ref{summary_photoshop} consecutively to $x_f$ and $x_b$ (i.e., $\Pi_{i=1}^{k_{f}}\varphi_{i}$ and $\Pi_{j=1}^{k_{b}}\phi_{j}$ in formula \ref{make_GreatSplicing}). 

\item Using the cut-out tools shown in Table \ref{summary_photoshop}, we extract a region from $x_f$ as the spliced region, paste it onto $x_b$ as a new layer, and save it in PNG format. If the spliced region is a recognizable semantic object, the spliced image is called an \textbf{object-awareness spliced image}. Conversely, if the spliced region is a random area without explicit content, the spliced image is called a \textbf{shape-random spliced image}. GreatSplicing comprises 2,887 object-awareness spliced images and 2,113 shape-random spliced images.

\item Utilizing the Color Overlay in blending options, we set spliced regions to white and background regions to black, generating groundtruths.
\end{enumerate}

\begin{figure}[t]
    \centering
    \includegraphics[scale=0.44]{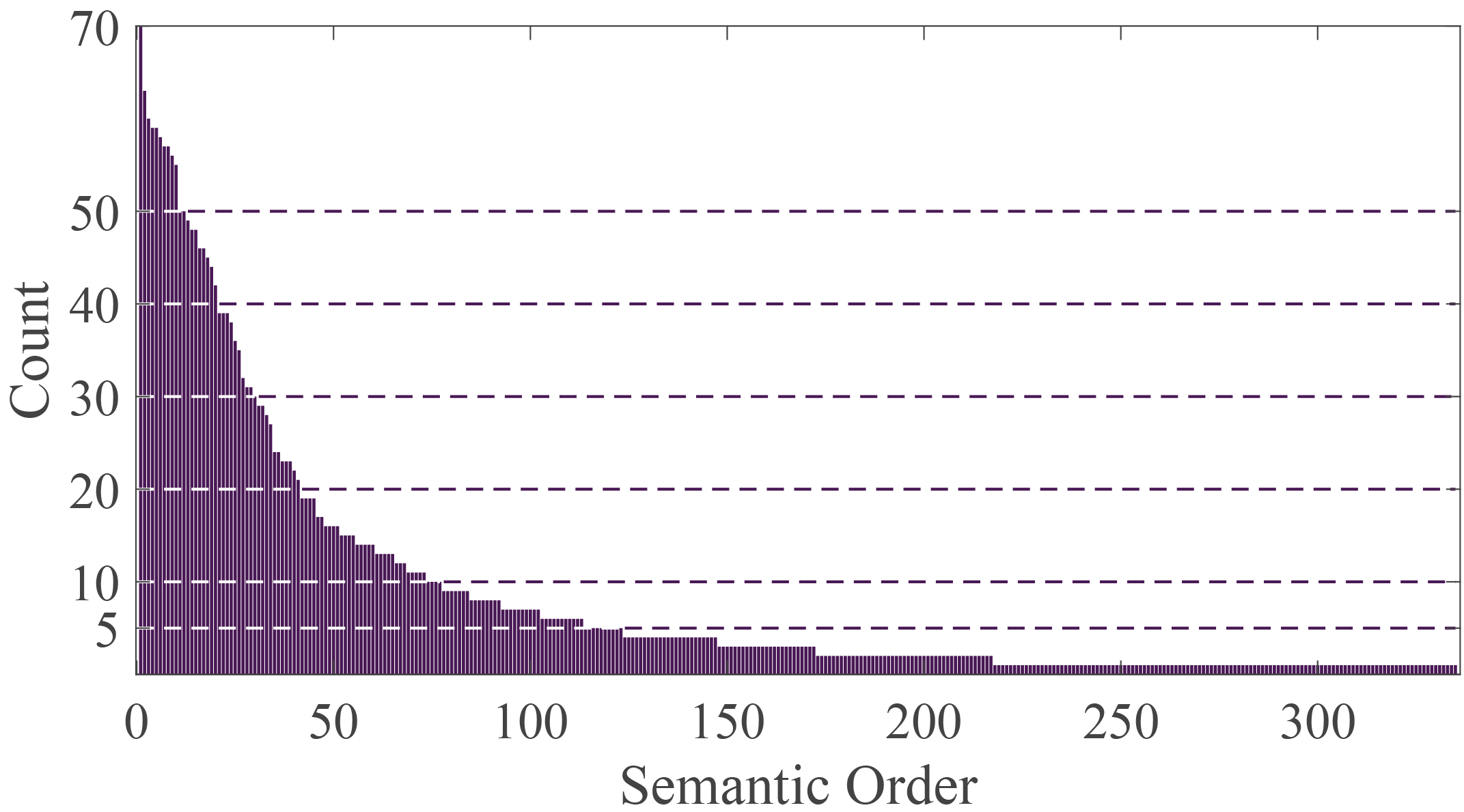}
    \caption{GreatSplicing comprises 335 distinct semantic categories for its spliced regions. In the distribution diagram, the horizontal axis represents the index of each semantic category, while the vertical axis represents the amount of spliced images corresponding to each semantic category, arranged in descending order.}
    \label{figure_semantic_nums_statistic}
\end{figure}

In addition to spliced images and their corresponding groundtruths, the following information for each spliced image is also recorded and compiled: 
\begin{enumerate}
\item Corresponding image index of $x_f$ and $x_b$ in BossBase;

\item Semantic content of the spliced regions;

\item Image manipulations undergone by $x_f$ and $x_b$;

\item Number of the connected spliced regions.

\item Annotation of the spliced region in $x_f$.
\end{enumerate}

GreatSplicing includes 335 distinct semantic categories for spliced regions, surpassing all existing splicing datasets. Figure \ref{figure_semantic_nums_statistic} presents a distribution diagram displaying the amount of spliced images for each semantic category. Additionally, the manual creation process of GreatSplicing ensures a high level of realism in the object-awareness spliced images. These images are comparable to those in real-world applications. Selected object-awareness spliced images from GreatSplicing are illustrated in Figure \ref{make_GreatSplicing}.

\section{Experiments}
This section provides extensive evaluations of GreatSplicing and existing splicing datasets based on state-of-the-art detection models. The empirical results demonstrate that GreatSplicing enables the detection models to learn splicing traces more effectively.

\subsection{Setup}
\noindent\textbf{Experimental environment.} All experiments are conducted on a single NVIDIA Tesla T4 GPU, with Ubuntu 18.04, Python 3.8, and PyTorch 1.9.1.

\begin{figure*}[t]
\centering
\includegraphics[scale=0.42]{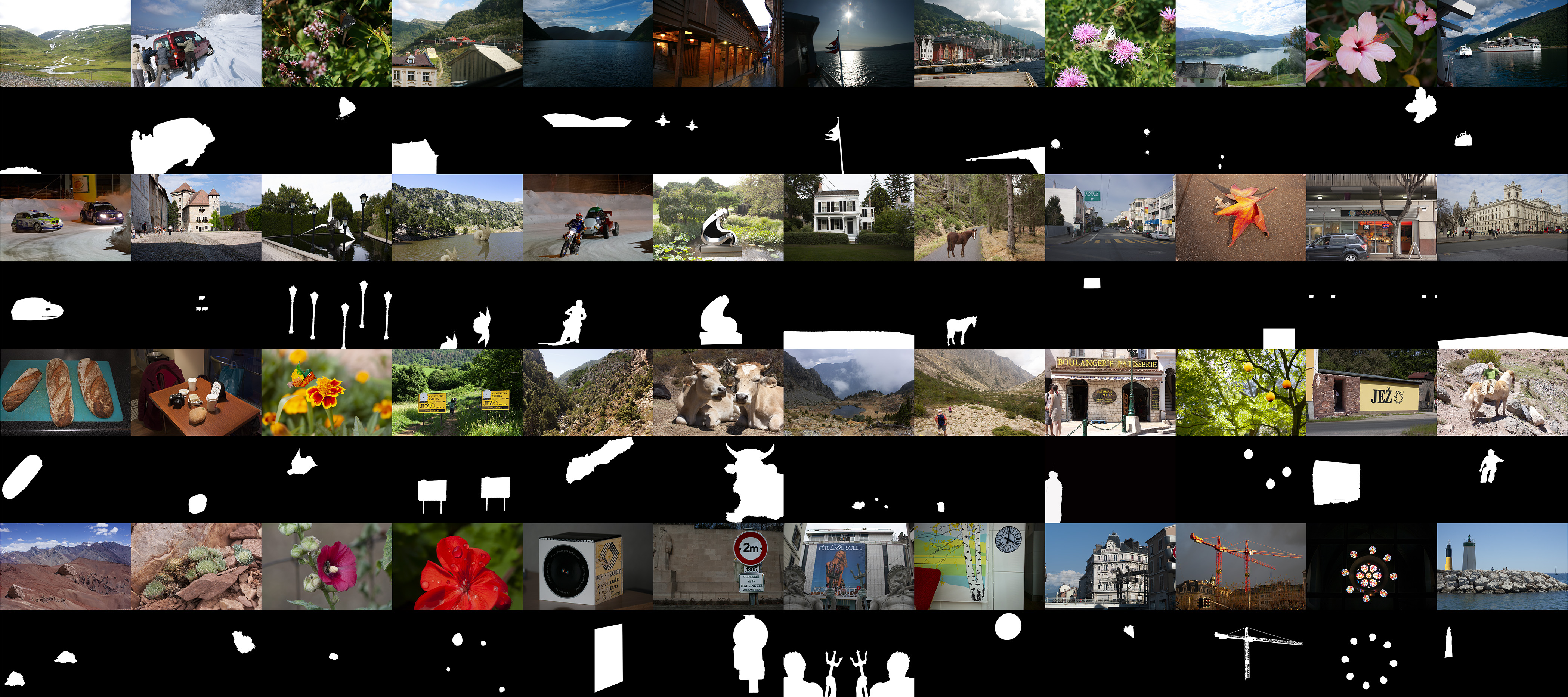}
\caption{Selected spliced images from GreatSplicing. GreatSplicing exhibits significant advantages, such as semantic richness, high verisimilitude, high resolution, etc.}
\label{method_greatsplicing}
\end{figure*}

\noindent\textbf{Datasets.} The existing splicing datasets used in the experiments include CASIA V2.0, DEFACTO, FantasticReality, NIST16, and Columbia. All images are resized to $448\times608$ and saved in PNG format. Specific processing methods for each dataset are detailed below:
\begin{itemize}
\item  CASIA-SP-1320 and CASIA-AU-700 are created by randomly selecting 1,320 spliced images and 700 authentic images from CASIA V2.0;
\item  DEFACTO-SP-3500 are formed by randomly selecting 500 images from each of seven semantic categories of spliced images;
\item FR-SP-3500 consists of 3,500 randomly selected spliced images from FantasticReality, while FR-AU-700 contains 700 authentic images randomly chosen from authentic images;
\item Columbia-SP-180 comprises all 180 spliced images from Columbia;
\item NIST16-SP-280 includes 280 spliced images from NIST16;
\item We randomly select 2,100 object-awareness spliced images and 1,400 shape-random spliced images from GreatSplicing to form GS-SP-3500. Additionally, we randomly select 300 object-awareness spliced images and 400 shape-random spliced images from the remaining GreatSplicing dataset to create GS-SP-700. Furthermore, 700 images are randomly selected from BossBase to create GS-AU-700.
\end{itemize}
Specifically, the selected authentic images have not been used as the source for spliced regions or background regions in the chosen spliced images.

\noindent\textbf{Models.} Three basic visual networks and two state-of-the-art splicing detection models are employed for evaluations. The three basic networks are VGG-16\cite{simonyan2014very}, ResNet-50\cite{he2016deep}, and U-Net\cite{ronneberger2015u}. We made minor adjustments to adapt them for end-to-end splicing detection. The two splicing detection models are RRU-Net\cite{bi2019rru} and MVSS-Net\cite{dong2022mvss}. They provide an objective and comprehensive reflection.

\subsection{Train-Test Validation}
To validate the effectiveness of GreatSplicing, a train-test evaluation is conducted. GS-SP-3500 is used as the training set, and GS-SP-700 is used as the testing set. These five models are trained and tested on GreatSplicing, with F1-score and IoU as the metrics. The results are depicted in Table \ref{experiment_table_train_test_validation}. The results show that segmentation-capable models, such as U-Net and RRU-Net, excel in learning splicing traces from GreatSplicing, leading to superior splicing detection performance. In contrast, standard visual convolutional networks like VGG-16 and ResNet-50, which lack specialized modules for splicing detection, perform subpar in train-test validation. 

It is worth noting that the relatively lower F1-score and IoU metrics in our experiments, compared to other research findings, can be attributed to the modest sample size of the 3,500 training samples we employed. Other splicing detection models are typically trained on substantially larger datasets achieved through synthesis or augmentation techniques. Furthermore, the comparative experiments in the existing works are often conducted on datasets with inherent semantic cues, enabling the models to identify the targets better. In contrast, GreatSplicing eliminates semantic interference, compelling the detection models to learn genuine splicing traces, thus increasing the learning complexity.

\begin{table}
    \centering
    \fontsize{9}{10}\selectfont
     \setlength\tabcolsep{1.5pt}
    \renewcommand{\arraystretch}{1.2}
    \begin{tabular}{cccccc}
        \toprule

        Model&VGG-16& U-Net& ResNet-50&MVSS-Net&RRU-Net\\
        \midrule
        F1&0.286&0.373&0.282&0.326&0.400\\ 
        IoU&0.216&0.305&0.221&0.267&0.340\\
        \bottomrule
    \end{tabular}
    \caption{Results of Train-Test Validation. Training set is GS-SP-3500 and testing set is GS-SP-700.}
    \label{experiment_table_train_test_validation}
\end{table}

\begin{table}[t]
    \centering
    \fontsize{9}{10}\selectfont
     \setlength\tabcolsep{0.5pt}
    \renewcommand{\arraystretch}{1.2}
    \begin{tabular}{cccccc}
        \toprule
        \multirow{2}{*}[-0.5ex]{Dataset}& \multicolumn{4}{c}{Model} \\        
        & VGG-16 & ResNet-50&U-Net&RRU-Net&MVSS-Net\\
        \midrule
        
        CASIA V2.0&0.046&0.070&0.023&\textbf{0.015}&0.012\\
        FantasticReality&0.067&0.112&0.082&0.083&0.063\\
        GreatSplicing&\textbf{0.007}&\textbf{0.008}&\textbf{0.005}&0.028&\textbf{0.008}\\
        \bottomrule
        
    \end{tabular}
    \caption{Results of authentic image misidentification. The criterion used in the table is MIR.}
    \label{experiment_table_authentic_images_misidentification}
\end{table}

\subsection{Authentic Image Misidentification}
Due to the limited semantic diversity in spliced regions, existing splicing datasets hinder omdels from effectively learning splicing traces. This leads to the extraction of irrelevant features and severe misidentification. In contrast, GreatSplicing offers diverse semantic categories in spliced regions, facilitating better splicing trace learning, reducing the extraction of irrelevant features, and lowering the misidentification of authentic images. To quantify misidentification, the misidentification rate (MIR) is defined as the ratio of the misidentification region's area to the authentic image's area, expressed as 
\begin{equation}
    MIR = \frac{FP}{P+N}.
\end{equation}

In this experiment, the CASIA and FantasticReality datasets are chosen because they offer authentic images, avoiding cross-domain confusion. Five models are trained on CASIA-SP-1320, FR-SP-3500, and GS-SP-3500, respectively, and are subsequently tested on CASIA-AU-700, FR-AU-700, and GS-AU-700, with MIR calculated. The results are shown in Table \ref{experiment_table_authentic_images_misidentification}. 
\begin{figure}[h]
    \centering
    \includegraphics[scale=0.48]{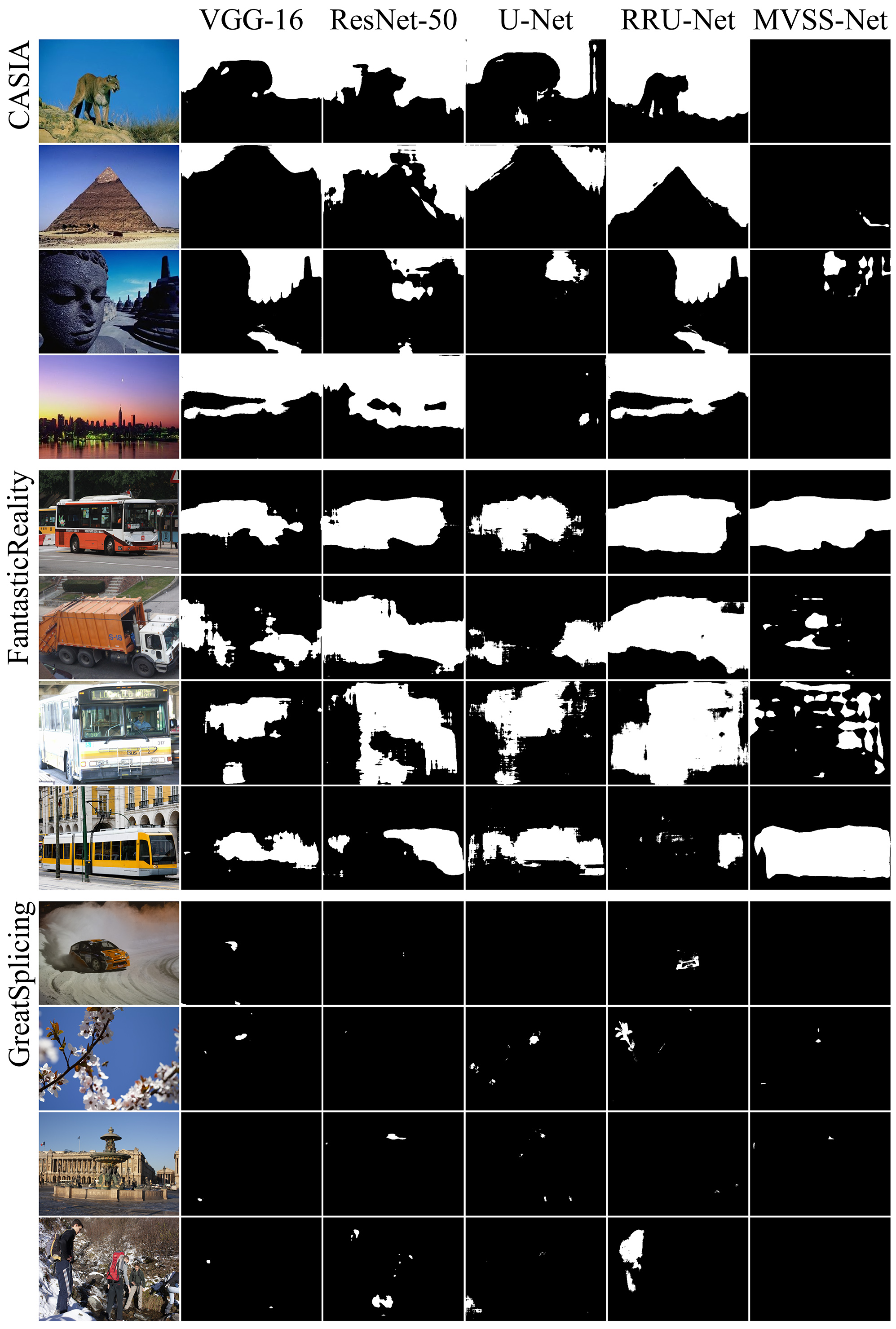}
    \caption{Results of authentic image misidentification. Twelve authentic color images were sourced from CASIA, FantasticReality, and BossBase. These images were tested with models trained on CASIA, FantasticReality, and GreatSplicing datasets respectively, resulting in misidentification outcomes.}
    \label{figure_experiment_misidentification}
\end{figure}
The results show that models trained on CASIA and FantasticReality have a high misidentification rate for authentic images. In contrast, models trained on GreatSplicing consistently perform the best, with misidentification rates approaching zero in most cases. 
\begin{table*}[t]
    \centering
    \fontsize{9}{10}\selectfont
     \setlength\tabcolsep{5.5pt}
    \renewcommand{\arraystretch}{1.2}
    \begin{tabular}{ccccccccccccc}
        \toprule
        \multirow{2}{*}[-0.5ex]{Test Set}    &  \multirow{2}{*}[-0.5ex]{Train Set}&  \multirow{2}{*}[-0.5ex]{Nums}    & \multicolumn{10}{c}{Model} \\ 
        
        & && \multicolumn{2}{c}{VGG-16} & \multicolumn{2}{c}{ResNet-50}&\multicolumn{2}{c}{U-Net}&\multicolumn{2}{c}{RRU-Net}&\multicolumn{2}{c}{MVSS-Net}\\
        \midrule

        &(Criterion)&&F1&IoU&F1&IoU&F1&IoU&F1&IoU&F1&IoU\\
        \midrule
        
        \multirow{4}{*}[-1.5ex]{Columbia}
        &CASIA V2.0&1,320&0.191&0.118&0.064&0.036&0.258&0.168&0.234&0.159&0.210&0.139\\
        &GreatSplicing&1,320&\textbf{0.226}&\textbf{0.138}&\textbf{0.204}&\textbf{0.128}&\textbf{0.292}&\textbf{0.194}&\textbf{0.390}&\textbf{0.298}&\textbf{0.301}&\textbf{0.209}\\
        \cmidrule{2-13}

        & FantasticReality&3,500&0.351&0.242&0.085&0.242&0.160&0.278&0.194&0.090&0.090
        &0.057\\
        &DEFACTO&3,500&\textbf{0.394}&\textbf{0.290}&0.006&0.003&0.240&0.157&0.062&0.037&0.019&0.011\\
        &GreatSplicing&3,500&0.282&0.185&\textbf{0.316}&\textbf{0.211}&\textbf{0.362}&\textbf{0.245}&\textbf{0.536}&\textbf{0.443}&\textbf{0.337}&\textbf{0.246}\\
        \midrule
        
        \multirow{4}{*}[-1.5ex]{NIST16}
        &CASIA V2.0&1,320&0.176&0.116&0.131&0.088&0.222&0.154&0.337&0.261&0.277&0.198\\
        &GreatSplicing&1,320&\textbf{0.280}&\textbf{0.198}&\textbf{0.314}&\textbf{0.228}&\textbf{0.331}&\textbf{0.248}&\textbf{0.411}&\textbf{0.318}&\textbf{0.336}&\textbf{0.254}\\
        \cmidrule{2-13}
        
        & FantasticReality&3,500&0.028&0.018&0.031&0.020&0.052&0.033&0.280&0.214&0.046&0.030\\
        & DEFACTO&3,500&0.302&0.204&0.064&0.048&0.341&0.238&0.261&0.199&0.070&0.049\\
        &GreatSplicing&3,500&\textbf{0.305}&\textbf{0.217}&\textbf{0.305}&\textbf{0.225}&\textbf{0.356}&\textbf{0.272}&\textbf{0.448}&\textbf{0.358}&\textbf{0.337}&\textbf{0.253}\\
        \bottomrule
        
    \end{tabular}
    \caption{Results of Cross-Dataset Splicing Detection. GreatSplicing exhibits optimal performance in nearly all test scenarios.}
    \label{Experiment_Cross_Dataset}
\end{table*}
\begin{figure*}[!t]
\centering
\includegraphics[scale=0.53]{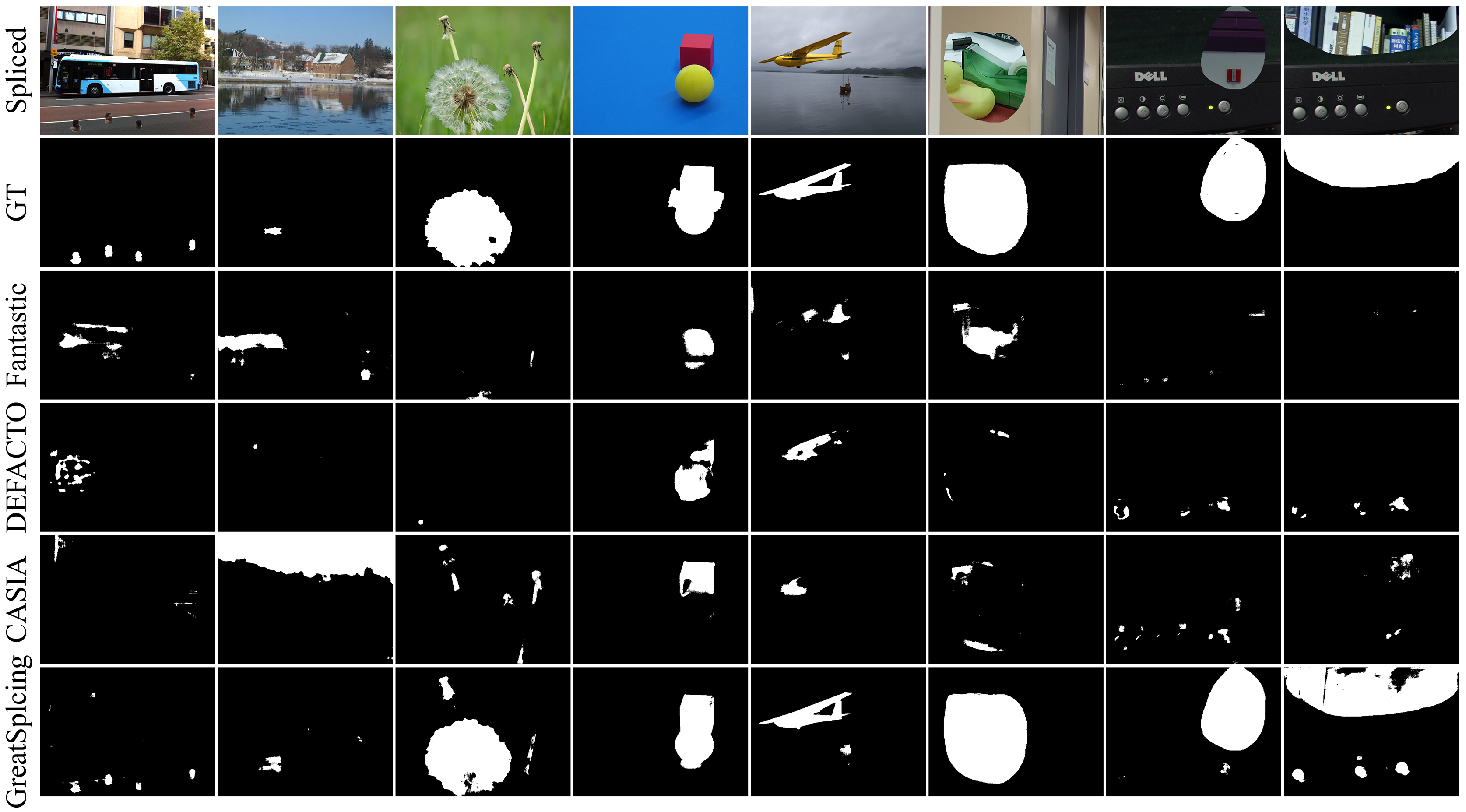}
\caption{Cross-Dataset Splicing Detection. We trained FantasticReality-SP-3500, DEFACTO-SP-3500, CASIA-SP-1320, and GreatSplicing-SP-3500 with RRU-Net and tested on NIST16 and Columbia.}
\label{figure_experiment_cross_dataset}
\end{figure*}

In addition, we display the misidentification results in Figure \ref{figure_experiment_misidentification}. The color images in the figure correspond to authentic images from CASIA, FantasticReality, and BossBase. Notably, CASIA-trained models show sensitivity to sky regions, while FantasticReality-trained models exhibit sensitivity to busses. Conversely, models trained on GreatSplicing display significantly reduced misidentification.

These facts show that datasets lacking diverse semantic categories in spliced regions cause the model to learn irrelevant features, leading to severe misidentification. GreatSplicing effectively helps detection models learn splicing traces, thus reducing the misidentification rate.

It is important to highlight that GreatSplicing does not trade off between misidentification and splicing detection rates. Instead, it achieves simultaneous improvement in both aspects. The following cross-dataset splicing detection experiments will provide additional clues.

\subsection{Cross-Dataset Splicing Detection}
This experiment aims to validate whether models trained on GreatSplicing exhibit superior generalization capabilities for splicing detection compared to existing datasets. We conduct cross-dataset experiments using CASIA-SP-1320, FantasticReality-SP-3500, DEFACTO-SP-3500, GreatSplicing-SP-1320, and GreatSplicing-SP-3500 as the training sets, and Columbia-SP-180 and NIST16-SP-280 as the test sets. The results are presented in Table \ref{Experiment_Cross_Dataset}, with the F1-score and IoU as the evaluation metrics. Selected visual results are presented in Figure \ref{figure_experiment_cross_dataset}.

The models trained on GreatSplicing exhibit consistently superior performance in cross-dataset splicing detection across almost all scenarios. This is attributed to the semantic diversity of GreatSplicing in the spliced regions, which enables the detection models to learn more general splicing traces, reducing reliance on irrelevant features. Besides, the cross-dataset experiment confirms that GreatSplicing achieves a dual benefit: reducing the misidentification rate while simultaneously enhancing the splicing detection rate, without a trade-off between the two.

\section{Conclusion}
The existing splicing datasets suffer from limited semantic diversity in the spliced regions, hindering models from learning general splicing traces effectively. This limitation leads to high misidentification rates in authentic images and decreased splicing detection capabilities. In addition, existing datasets exhibit a variety of flaws stemming from improper construction. To address these issues, our GreatSplicing offers rich semantic diversity in the spliced regions, encompassing 335 semantic categories. Furthermore, GreatSplicing follows standardized production procedures, overcoming the deficiencies present in existing splicing datasets. Extensive experiments confirm that models trained on GreatSplicing significantly reduce misidentification rates while enhancing splicing detection performance, simultaneously improving both aspects.

\bibliography{aaai2026}

\end{document}